\documentclass[conference]{IEEEtran}
\IEEEoverridecommandlockouts
\usepackage{spconf,amsmath,epsfig}

\let\OLDthebibliography\thebibliography
\renewcommand\thebibliography[1]{
  \OLDthebibliography{#1}
  \setlength{\parskip}{0pt}
  \setlength{\itemsep}{0pt plus 0.3ex}
}
\usepackage{color}
\usepackage{booktabs}
\usepackage{arydshln}
\usepackage{multirow,subfigure}
\usepackage{hyperref,MnSymbol}
\begin{document}\sloppy

\def\x{{\mathbf x}}
\def\L{{\cal L}}

\title{DDH-QA: A Dynamic Digital Humans Quality Assessment Database}
%

\name{Zicheng Zhang$\tiny{^{1,2}}$,Yingjie Zhou$\tiny{^{1,2}}$,Wei Sun$\tiny{^{1,2}}$,Wei Lu$\tiny{^{1,2}}$,Xiongkuo Min$\tiny{^{1,2}}$,Yu Wang$^{1}$,and Guangtao Zhai$\tiny{^{1,2,3}}$}
\address{ $^{1}$Institute of Image Communication and Network Engineering, Shanghai Jiao Tong University, China \\
 $^{2}$ Peng Cheng Laboratory, China \\
$^{3}$ MoE Key Lab of Artificial Intelligence, AI Institute, Shanghai Jiao Tong University, China
\thanks{This work was supported in part by NSFC (No.62225112, No.61831015), the Fundamental Research Funds for the Central Universities, National Key R\&D Program of China 2021YFE0206700, and Shanghai Municipal Science and Technology Major Project (2021SHZDZX0102).}}

\maketitle

\begin{abstract}
In recent years, large amounts of effort have been put into pushing forward the real-world application of dynamic digital human (DDH). However, most current quality assessment research focuses on evaluating static 3D models and usually ignores motion distortions. Therefore, in this paper, we construct a large-scale dynamic digital human quality assessment (DDH-QA) database with diverse motion content as well as multiple distortions to comprehensively study the perceptual quality of DDHs. Both model-based distortion (noise, compression) and motion-based distortion (binding error, motion unnaturalness) are taken into consideration.
Ten types of common motion are employed to drive the DDHs and a total of 800 DDHs are generated in the end. Afterward, we render the video sequences of the distorted DDHs as the evaluation media and carry out a well-controlled subjective experiment. Then a benchmark experiment is conducted with the state-of-the-art video quality assessment (VQA) methods and the experimental results show that existing VQA methods are limited in assessing the perceptual loss of DDHs. The database is available at https://github.com/zzc-1998/DDH-QA.
\end{abstract}
\begin{keywords}
Dynamic digital human, model-based distortion, motion-based distortion, subjective experiment
\end{keywords}

\section{Introduction}
\begin{figure}[!t]
    \centering
    \includegraphics[width=\linewidth]{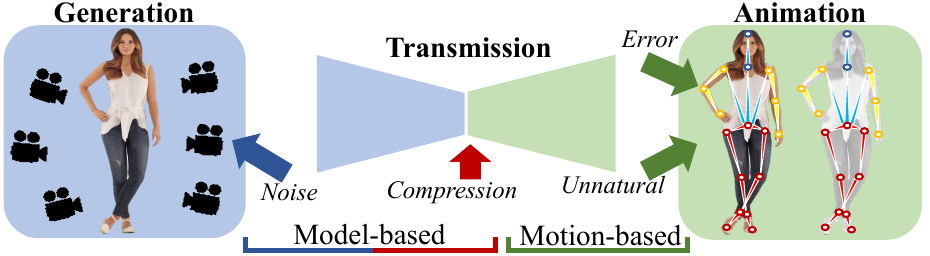}
    \caption{Distortion sources of DDHs.}
    \label{fig:source}
\end{figure}

Digital humans indicate digital models represented by computer graphics, which are usually static and fixed. Dynamic digital humans (DDHs) are digital models driven by predefined animations \cite{kim2006prediction}, which have been widely adopted in applications such as the game industry, film post-production, metaverse, etc.  
As shown in Fig. \ref{fig:source}, the DDHs suffer from both model-based and motion-based distortions. The model-based distortions represent the degradations directly affecting the digital human models. For example, the slight geometry shift and inevitable camera noise can introduce noise disturbance to the geometry structure and texture maps during the generation procedure \cite{zhang2022treating,zhang2022mm}. Moreover, to support real-time VR/AR applications under restricted bandwidth, the digital human models always undergo compression through the transmission procedure. The motion-based distortions stand for the incoherence and unnaturalness of the animation, which are often caused by inappropriate skeleton binding and confusing motion.  
Nowadays, researchers are mainly paying attention to the generation, representation, rendering, and animation of digital humans \cite{zhu2019applications}. However, the quality assessment for DDHs has fallen behind and effective approaches along with databases are urgently needed. 

Therefore, we propose the first dynamic digital human quality assessment (DDH-QA) to tackle the mentioned challenge. One male and one female digital humans represented by texture meshes are collected as the reference. Then we introduce six model-based distortions (color noise, geometry noise, texture compression, texture downsampling, position compression, and UV map compression) and two motion-based distortions (skeleton binding error and motion range unnaturalness) to reference models, which generate 800 distorted DDHs in total. Afterward, we carry out a  subjective experiment to collect the perceived quality ratings for the distorted DDHs. Finally, we conduct a benchmark experiment on the DDH-QA database with state-of-the-art video quality assessment (VQA) methods, which shows that current quality assessment methods are not effective for predicting the visual quality levels of DDHs. Our contributions can be summarized as followed:
\begin{itemize}
    \item\textbf{To the best of our knowledge, we construct the first dynamic digital human quality assessment database}, which provides 800 distorted DDHs with both model-based and motion-based distortions. 
    \item We carry out a subjective study to collect the perceptual quality labels of the distorted DDHs. A total of 32,800 = 41$\times$800 quality ratings are gathered.
    \item We conduct a benchmark experiment to exhibit the performance of the existing state-of-the-art quality assessment methods.
\end{itemize}

\begin{table}[!htp] \small
\centering
\renewcommand\arraystretch{0.9}
\caption{The comparison of 3D-QA databases and our database, where `Num' represents the number of the models provided with quality labels.}
\begin{tabular}{l|c|c}
\toprule
Database       &Num  &Content              \\
\midrule
SJTU-PCQA \cite{sjtu-pcqa}    &378  & Colored Point Cloud      \\
WPC \cite{liu2022perceptual}  &740  &  Colored Point Cloud   \\
LSPCQA \cite{liu2022point}    &1,240    &Colored Point Cloud   \\
CMDM \cite{cmdm}              &80  & Colored Mesh    \\
TMQA \cite{nehme2022textured} &3,000 & Textured Mesh\\
VVDB2 \cite{zerman2020textured}        &152 & Volumetric Video \\
DHHQA \cite{zhang2022perceptual}      & 1,540 & Static Digital Human Head  \\
\textbf{DDH-QA}(Ours) & 800 & Dynamic Digital Human\\
\bottomrule
\end{tabular}
\label{tab:comparison}
\end{table}

\section{Related Works}
\subsection{3D Model Quality Assessment Database}
In this section, we give a brief review of the 3D model quality assessment (3D-QA) databases.  
Mainstream 3D-QA databases focus on static point cloud quality assessment (PCQA) and mesh quality assessment (MQA) \cite{zhang2021mesh,fan2022no,zhang2022no}. Namely, the SJTU-PCQA \cite{sjtu-pcqa}, WPC \cite{liu2022perceptual}, and LSPCQA \cite{liu2022point} databases contain 378, 740, and 1,240 subjective annotated colored point clouds respectively, which are distorted by noise, downsampling, and compression. Some researchers are also interested in MQA tasks. For example, the CMDM \cite{cmdm} database employs simplification and quantization algorithms to obtain 80 distorted colored meshes. The TMQ database further provides 3,000 distorted textured meshes by compressing both the geometry structure and texture maps. All the databases mentioned above are constructed for common 3D objects, then some databases are proposed to focus on 3D digital humans. The VVDB2 \cite{zerman2020textured} database provides 152 3D human volumetric videos and the DHHQA \cite{zhang2022perceptual} database includes 1,540 distorted static digital human heads. However, none of these databases specifically investigate the perceptual quality of DDHs and all of them ignore the motion distortion.

\subsection{VQA Development}
Since most DDHs are presented in the format of rendered 2D videos, it is reasonable to transfer the VQA models to the DDH-QA tasks.  The VQA methods can be generally categorized into full-reference (FR) and no-reference (NR) VQA methods according to the availability of the reference videos. The FR-VQA methods usually compare the frame-level difference with the assistance of image quality assessment (IQA) metrics such as PSNR and SSIM \cite{ssim}. For the NR-VQA development, some handcrafted-based methods \cite{brisque,saad2014blind,korhonen2019two,tu2021ugc} are proposed to extract features based on natural scene statistics (NSS) and regress the features via the Support Vector Machine. With the development of deep neural networks (DNN), some researchers \cite{tu2021rapique,sun2022deep,wu2022fast} propose to utilize DNNs for feature extraction and have greatly boosted the performance of NR-VQA models.

\section{Database Construction}

\begin{figure}[!t]
    \centering
    \subfigure[Male]{\begin{minipage}[t]{0.4\linewidth}
                \centering
                \includegraphics[height = 4cm]{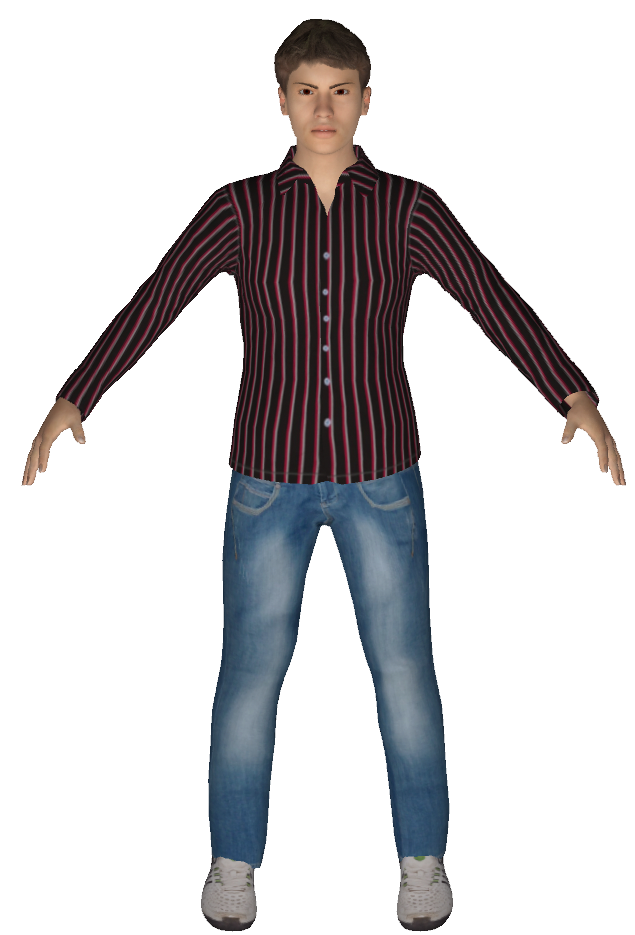}
                \end{minipage}}
    \subfigure[Female]{\begin{minipage}[t]{0.58\linewidth}
                \centering
                \includegraphics[height = 4cm]{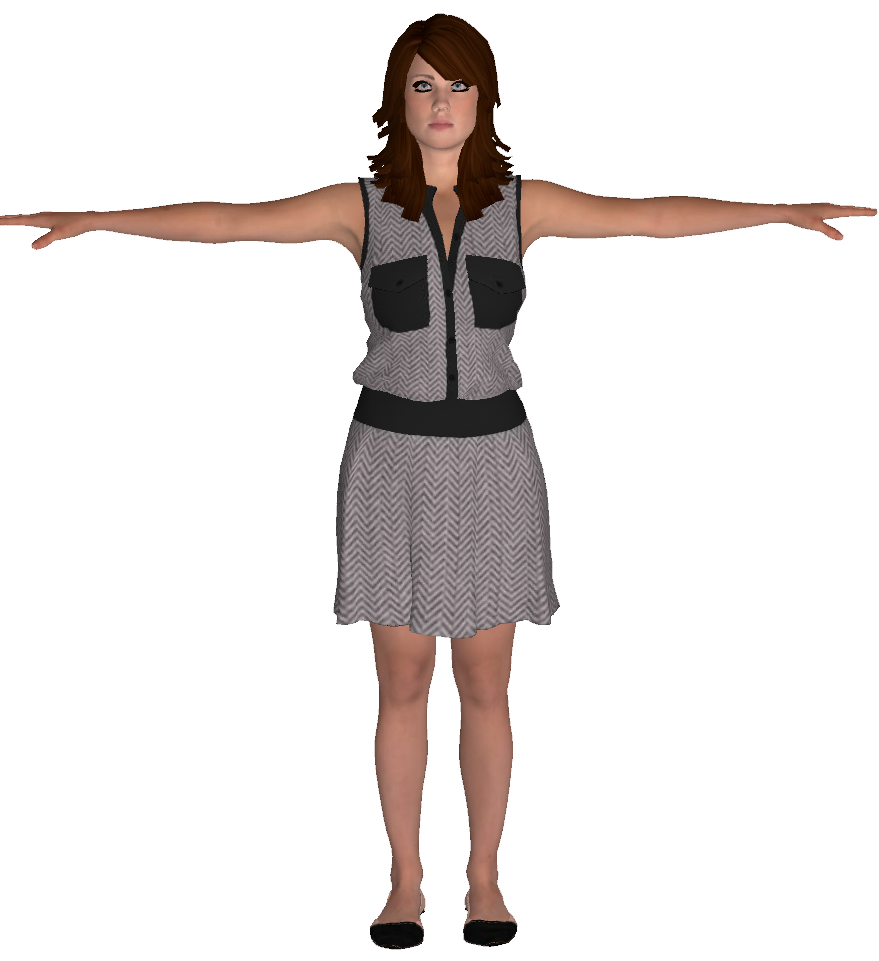}
                \end{minipage}}
    \caption{Illustration of the source male and female digital human models. The male digital human model is displayed in `A' pose while the female digital human model is displayed in `T' pose respectively.}
    \label{fig:model}
\end{figure}

\subsection{Source Model Collection}
To build the dynamic digital human quality assessment (DDH-QA) database, we collect one male and one female digital human models which can be freely downloaded from \textbf{Cgtrader}\footnote{https://www.cgtrader.com/} and are represented by textured meshes.  The male model contains 19,528 vertices and 2K texture maps while the female model contains 16,351 vertices and 2K texture maps. The front projections of the digital human models are illustrated in Fig. \ref{fig:model}.

\subsection{Distortion Generation}
To fit the practical situation of producing DDHs, we introduce two types of distortions, including model-based and motion-based distortions. The overview of the introduced distortion is shown in Table \ref{tab:distortion}.

\begin{figure}[!t]
    \centering
    \subfigure[CN]{\begin{minipage}[t]{0.32\linewidth}
                \centering
                \includegraphics[width=0.751\linewidth]{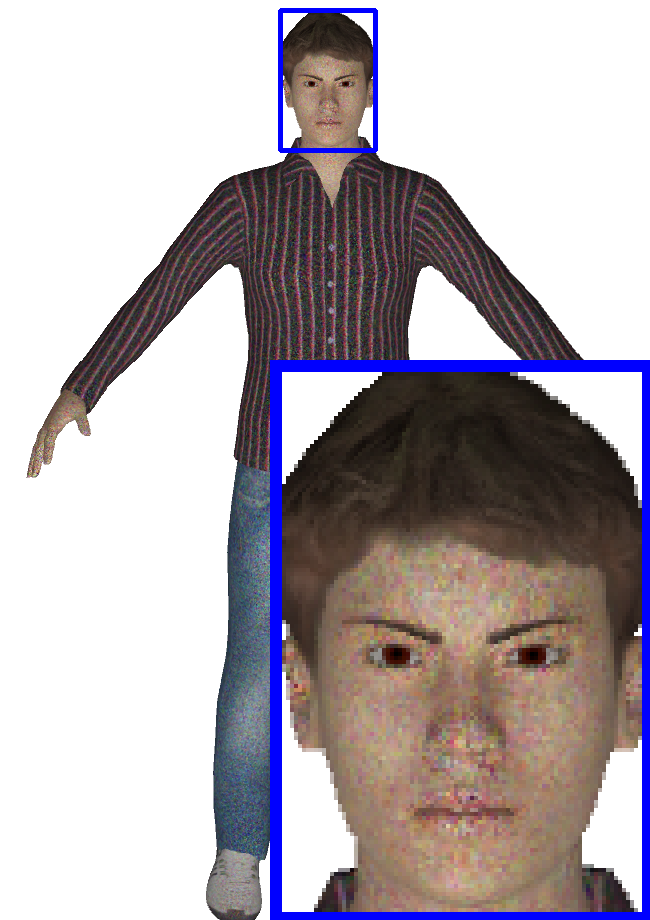}
                \end{minipage}}
    \subfigure[GN]{\begin{minipage}[t]{0.32\linewidth}
                \centering
                \includegraphics[width=0.751\linewidth]{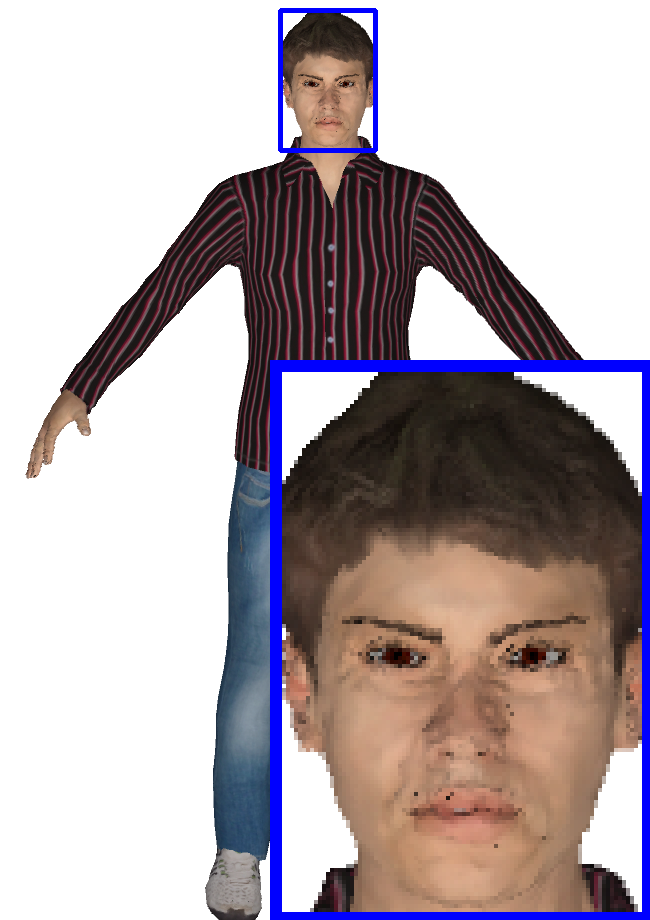}
                \end{minipage}}
    \subfigure[TC]{\begin{minipage}[t]{0.32\linewidth}
                \centering
                \includegraphics[width=0.751\linewidth]{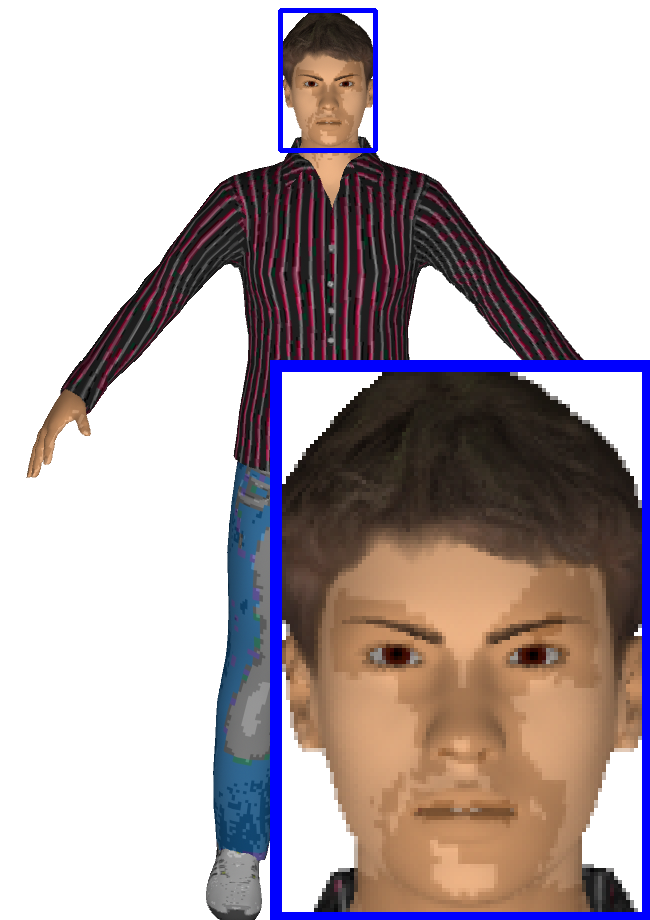}
                \end{minipage}}
    \subfigure[TD]{\begin{minipage}[t]{0.32\linewidth}
                \centering
                \includegraphics[width=0.751\linewidth]{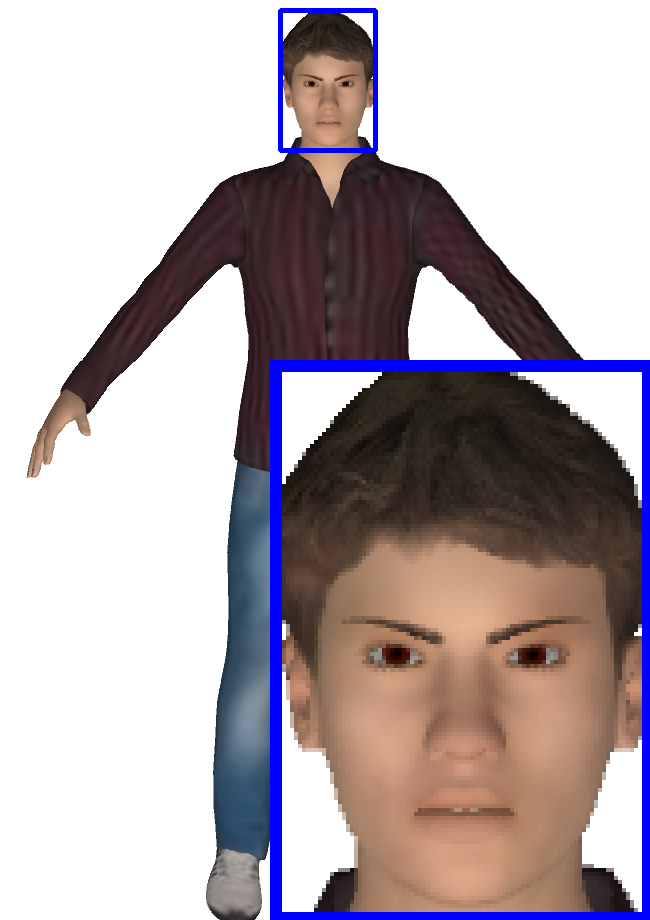}
                \end{minipage}}
    \subfigure[PC]{\begin{minipage}[t]{0.32\linewidth}
                \centering
                \includegraphics[width=0.751\linewidth]{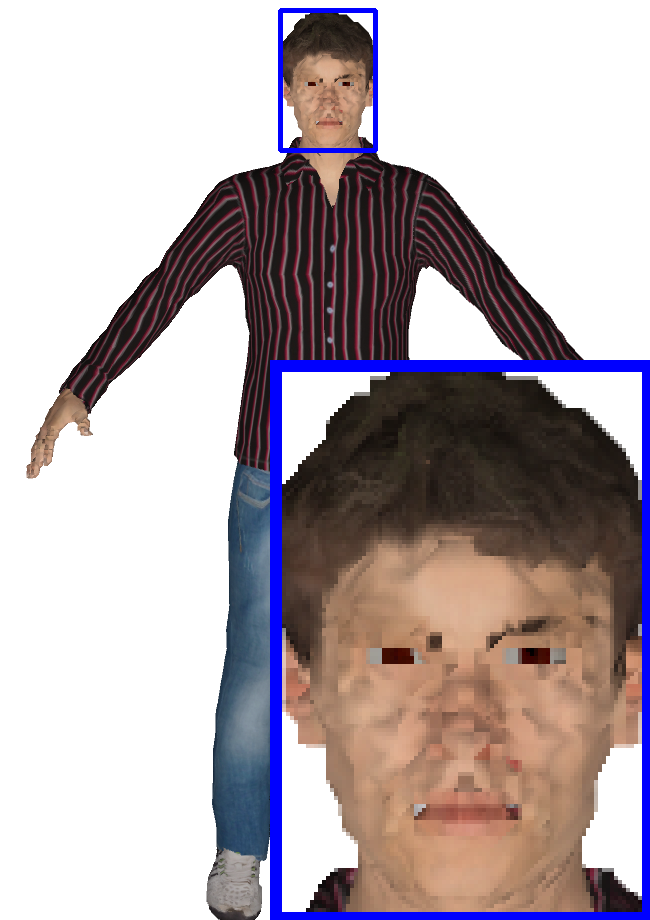}
                \end{minipage}}
    \subfigure[UMC]{\begin{minipage}[t]{0.32\linewidth}
                \centering
                \includegraphics[width=0.751\linewidth]{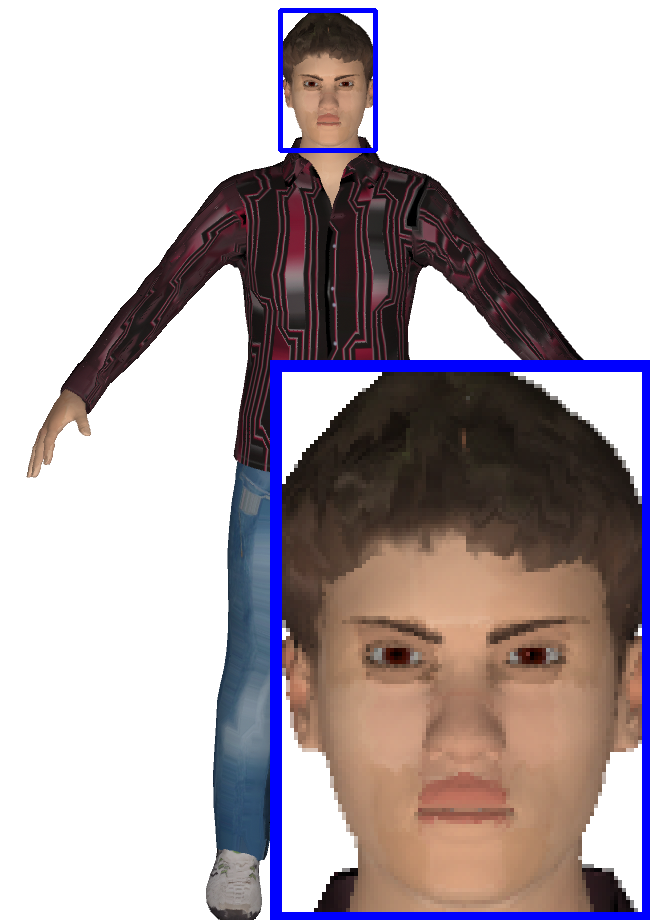}
                \end{minipage}}
    \caption{Projection samples of the model-based distortions, from which we can see that different kinds of distortions can cause diverse perceptual loss to the digital human models.}
    \label{fig:model_distortion}
\end{figure}

\textbf{(1) Model-based Distortion}:
The model-based distortions focus on the noise and compression artifacts during the generation and transmission procedure, which include: (a) Color Noise (CN): Gaussian noise is added to the texture maps with $\sigma_{c}$ set as \{20, 40, 60, 80, 100\}; (b) Geometry Noise (GN): Gaussian noise is introduced to the vertices with $\sigma_{g}$ set as \{0.01, 0.02, 0.03, 0.04, 0.05\}; (c) Texture Compression (TC): The texture maps are compressed with JPEG and the quality levels are set as \{3, 7, 15, 20, 25\}; (d) Texture Downsampling (TD): The texture maps are downsampled with the sampling rate of \{2, 4, 8, 12, 16\}; (e) Position Compression (PC): The position attributes are quantified wit the Draco \cite{Draco} library and quantization parameters {\bf qp} are set as \{6, 7, 8, 9, 10\};
(f) UV Map Compression (UMC): The texture coordinate attributes are quantified with the Draco library and the quantization parameters {\bf qt} are set as \{3, 4, 5, 6, 7\}; The distorted samples are exhibited in Fig. \ref{fig:model_distortion}

\begin{table}[!t] \small
\renewcommand\tabcolsep{4.2pt}
\setlength{\abovecaptionskip}{-5pt}
  \caption{Overview of the introduced distortion.}
  \begin{center}
  \begin{tabular}{c|c|cccc}
    \toprule
    Type &  Distortion & \multicolumn{4}{c}{Description} \\ \hline
    \multirow{6}{30pt}{Model-based} 
    &CN  &\multicolumn{4}{c}{Color noise on the texture maps}\\
    &GN  &\multicolumn{4}{c}{Geometry noise on the vertices}\\
    &TC  &\multicolumn{4}{c}{Texture maps JPEG compression}\\
    &TD  &\multicolumn{4}{c}{Texture maps downsampling}\\
    &PC  &\multicolumn{4}{c}{Position quantization by Draco}\\
    &UMC  &\multicolumn{4}{c}{Texture coordinate quantization by Draco}\\\hdashline
    \multirow{2}{30pt}{Motion-based} 
    &SBE &\multicolumn{4}{c}{Skeleton binding error}\\
    &MRU &\multicolumn{4}{c}{Unnatural motion range}\\
    \bottomrule
  \end{tabular}
  \end{center}
  \label{tab:distortion}

\end{table}

\begin{figure}[!tbp]
    \centering
    \subfigure[SBE]{\begin{minipage}[t]{0.49\linewidth}
                \centering
                \includegraphics[height=.85\linewidth]{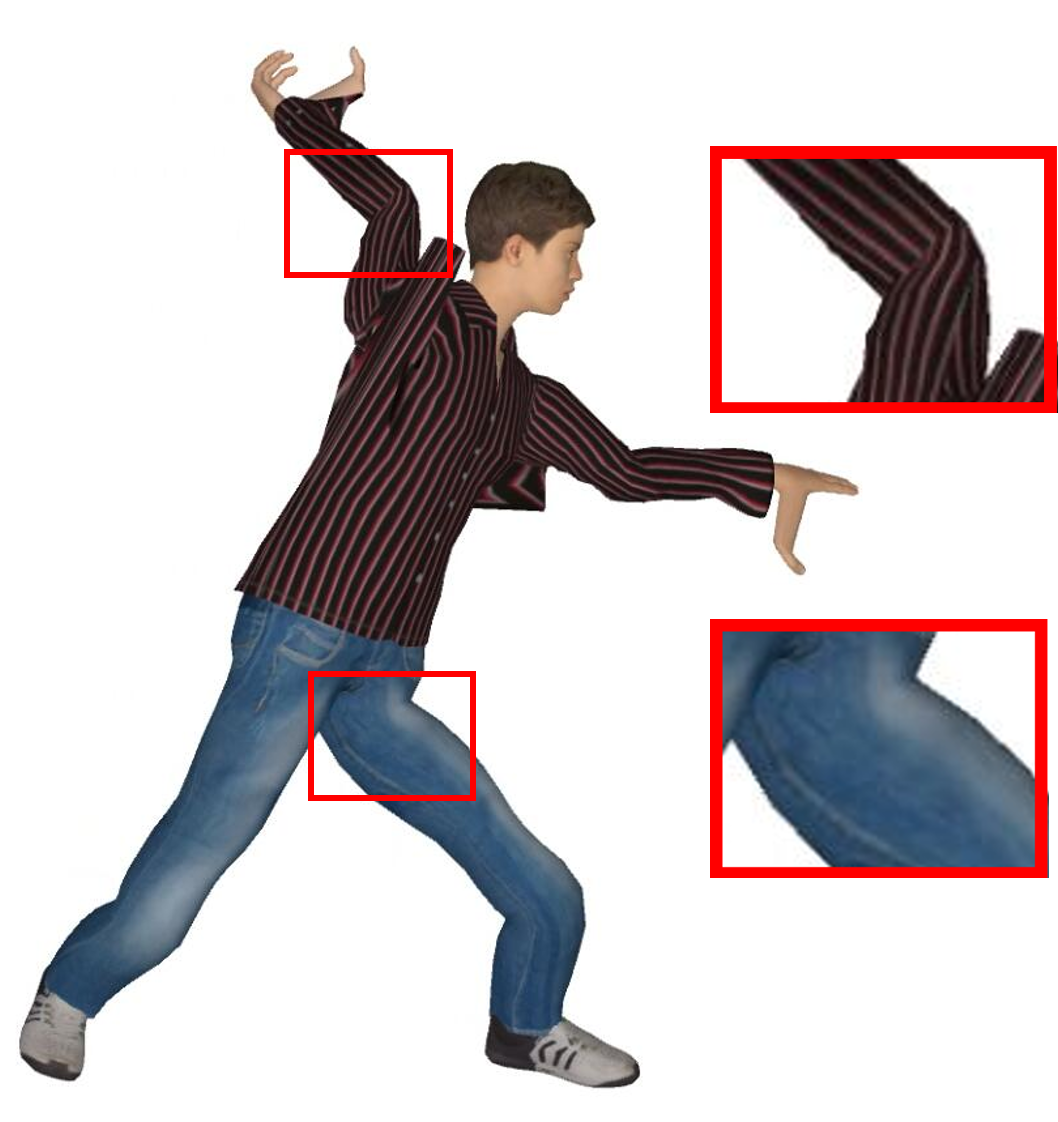}
                \end{minipage}}
    \subfigure[MRU]{\begin{minipage}[t]{0.49\linewidth}
                \centering
                \includegraphics[height=.85\linewidth]{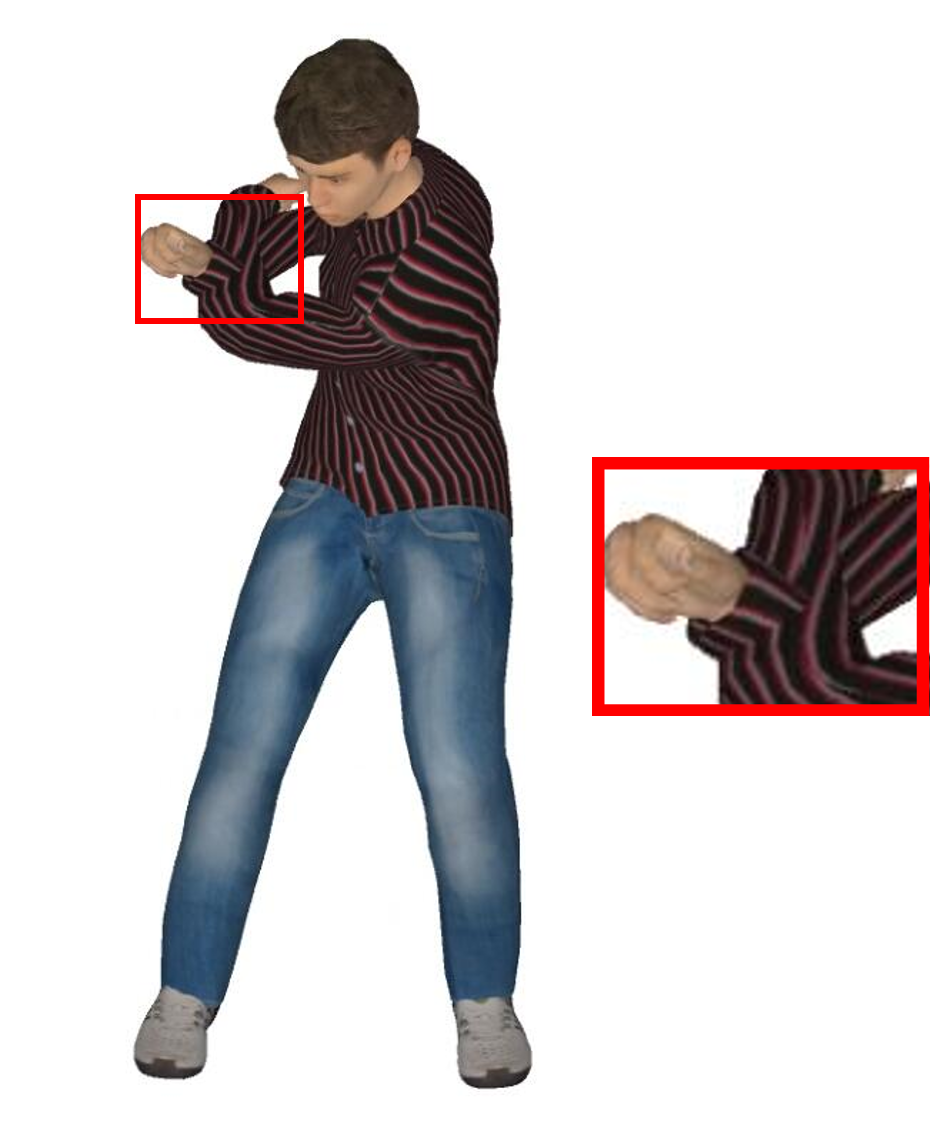}
                \end{minipage}}
    \caption{Examples of the motion-based distortions. The SBE distortions can severely discord the digital human body and rigidly twist the body joints. The MRU distortions usually cause model clipping, which makes the motion awkward and unnatural. }
    \label{fig:motion_distortion}
\end{figure}

\textbf{(2) Motion-based Distortion:}
The motion-based distortions focus on the skeleton rigging bias along with the motion unnaturalness.
In most piratical situations, digital humans are first processed with skeleton binding and then animated with the designed motion. The skeleton binding error and the poor-designed motion can cause confusion and unnaturalness to the dynamic digital human animation. Therefore, we introduce the motion-based distortions from two aspects: (a) Skeleton Binding Error (SBE) : Mismatch of skeleton key points are added under five manual defined levels to cover most quality ranges (slight mismatch $\sim$ severe mismatch); (b) Motion Range Unnaturalness (MRU) : We manually adjust the motion range to model the motion unnaturalness under five strengthens. The examples of the motion-based distortions are shown in Fig. \ref{fig:motion_distortion}.

\subsection{Video Rendering}
\begin{figure}[!t]
    \centering
    \includegraphics[width = \linewidth]{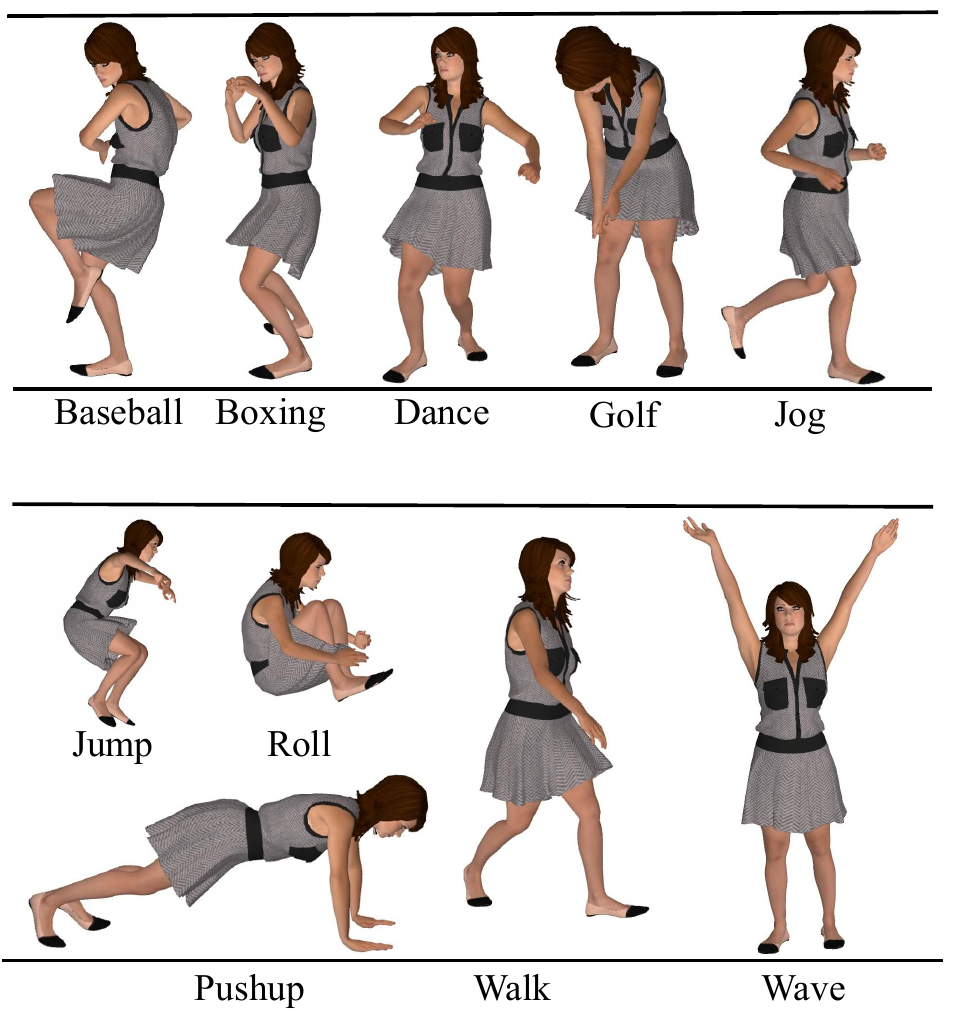}
    \caption{Illustration of the selected kinds of motion. Simple daily activities such as walk and wave are included. Complicated sports such as baseball and gold are considered as well.}
    \label{fig:motion}
\end{figure}

Since the DDHs are usually perceived in the format of 2D animation videos, we decide to render the DDHs into videos for evaluation.
We bind the skeleton of the digital human models and render the animation videos with a resolution of 1080P by using Maya software \footnote{https://www.autodesk.com/products/maya/} (the viewpoints are manually selected to cover sufficient quality content). To enrich the motion content diversity, we select ten types of common motion, including baseball, boxing, dance, golf, jog, jump, pushup, roll, walk, and wave. The overview of the ten kinds of motion is exhibited in Fig. \ref{fig:motion}. To sum up, a total of 800 = 2$\times$8$\times$5$\times$10 (digital human models$\times$distortion types$\times$distortion levels$\times$motion types) DDH video sequences are generated for evaluation.

\begin{figure}[!t]
    \centering
    \includegraphics[width=.9\linewidth]{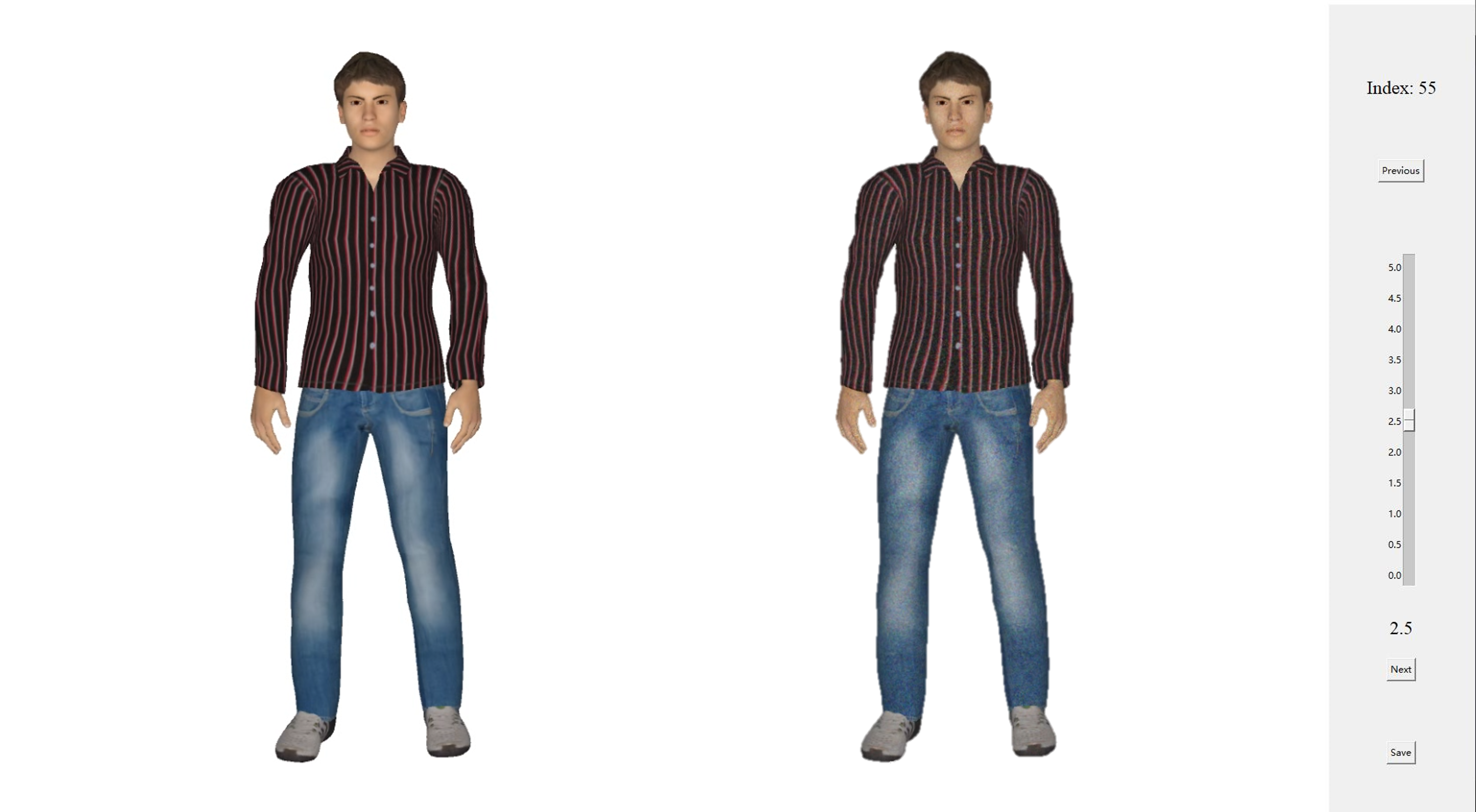}
    \caption{The screenshot of the subjective quality assessment interface. The reference videos (left) and the distorted videos (right) are displayed at the same time. }
    \label{fig:screen}
\end{figure}

\subsection{Subjective Experiment}
We carry out the subjective quality assessment experiment in a well-controlled laboratory environment under the instructions of ITU-R BT.500-13 \cite{bt2002methodology}. The rendered distorted dynamic digital human videos along with the corresponding reference ones are randomly displayed on a customized graphical subjective quality assessment interface, whose screenshot is shown in Fig. \ref{fig:screen}. We employ an iMac monitor for display, which supports a resolution up to 4096 $\times$ 2304.

A total of 41 subjects (20 males and 21 females) are invited to participate in the subjective experiment. All the subjects are seated from a distance of twice the screen height to the screen in an indoor environment with normal illumination levels. Before the subjective experiment, an instruction session is performed to help the subjects get familiar with the quality assessment task. The whole subjective experiment is split into 16 sessions and each session contains 50 distorted DDH video sequences. There is a 30-minutes break between each session and each subject is allowed to attend no more than 4 sessions in a single day. During each session, the distorted video sequence is played only once and the participants can rate the DDH quality according to the rendered DDH video from 1 to 5, with a minimum interval of 0.1. We ensure that each distorted DDH video is evaluated by the 41 invited participants and 32,800=800$\times$41 subjective ratings are collected in all.

\begin{figure}[!t]
    \centering
    \subfigure[]{\begin{minipage}[t]{\linewidth}
                \centering
                \includegraphics[width=0.9\linewidth]{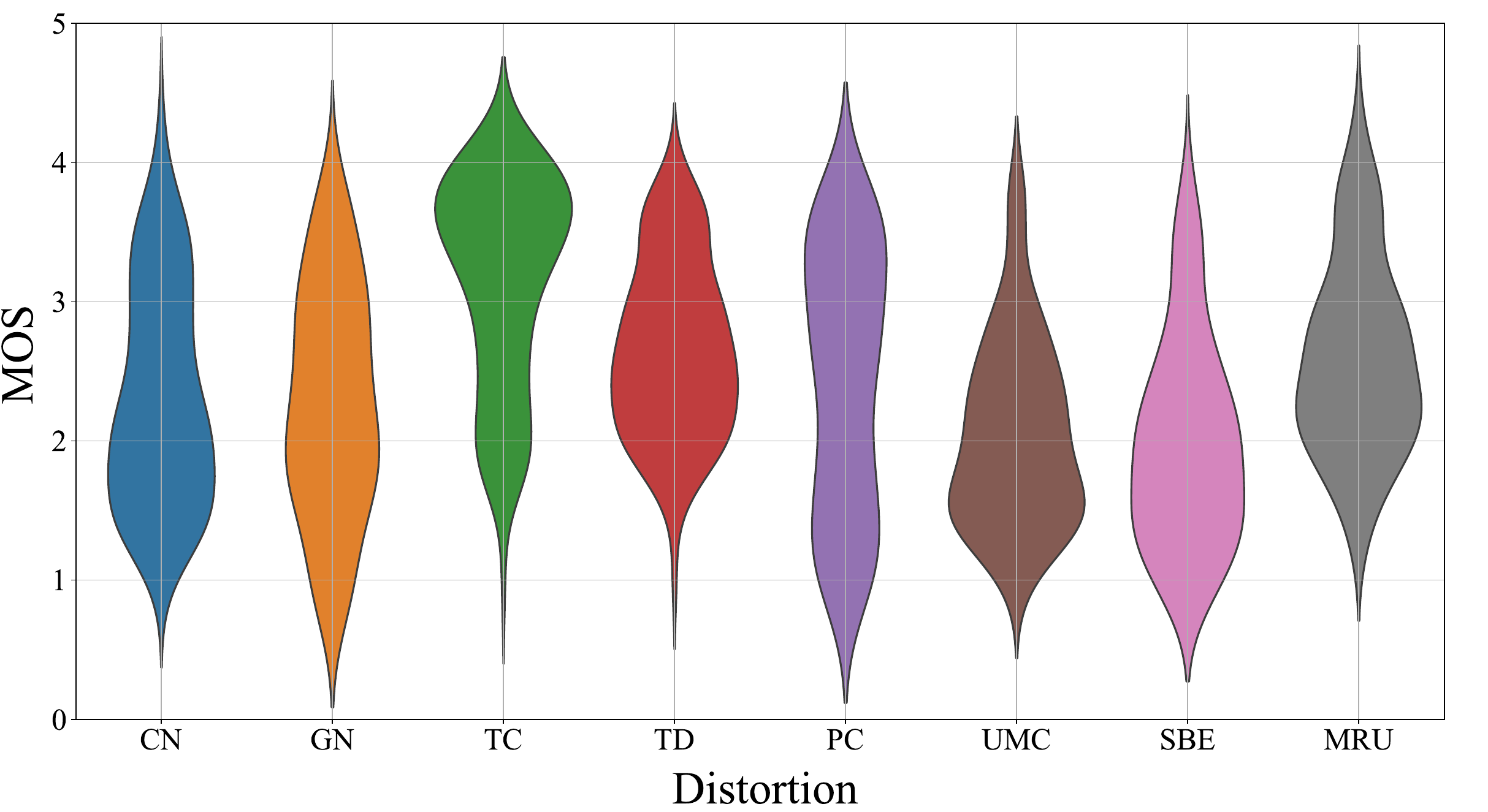}
                \end{minipage}} 
    \subfigure[]{\begin{minipage}[t]{\linewidth}
                \centering
                \includegraphics[width=0.9\linewidth]{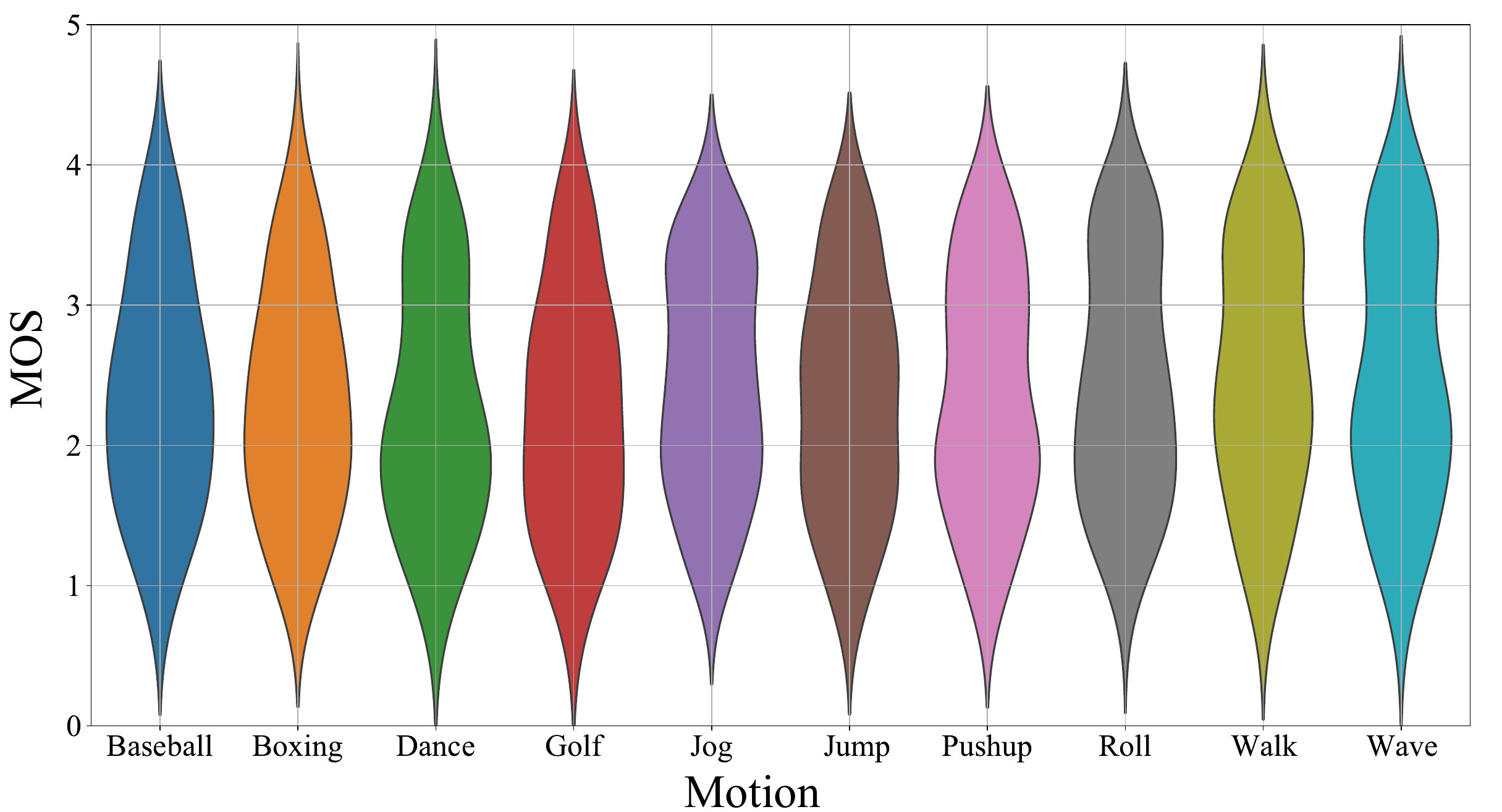}
                \end{minipage}}

    \caption{Distributions of the MOSs corresponding to the distortion and motion types. }
    \label{fig:distribution}
\end{figure}

\subsection{Subjective Data Processing}
Following the recommendation of ITU-R BT.500-13 \cite{bt2002methodology}, we calculate the z-scores as the quality labels of corresponding DDHs:

\begin{equation}
z_{i j}=\frac{r_{i j}-\mu_{i}}{\sigma_{i}},
\end{equation}
where $r_{ij}$ represents the quality rating given by the $i$-th subject on the $j$-th DDH, $\mu_{i}=\frac{1}{N_{i}} \sum_{j=1}^{N_{i}} r_{i j}$, $\sigma_{i}=\sqrt{\frac{1}{N_{i}-1} \sum_{j=1}^{N_{i}}\left(r_{i j}-\mu_{i}\right)}$, and $N_i$ is the number of DDH assessed by subject $i$. Additionally, we reject the quality ratings from unreliable subjects with the recommended subject rejection procedure proposed in \cite{bt2002methodology}.
In the end, the z-scores are linearly rescaled to $[1,5]$ and the mean opinion score (MOS) of DDH $j$ is obtained by averaging the rescaled z-scores: 

\begin{equation}
M O S_{j}=\frac{1}{M} \sum_{i=1}^{M} z_{i j}^{'},
\end{equation}
where $M O S_{j}$ represents the MOS for the $j$-th DDH, $M$ is the number of the valid subjects, and $z_{i j}^{'}$ are the rescaled z-scores.

\subsection{Subjective Data Analysis}
We further plot the distributions of the MOSs from the perspective of distortion and motion types, which are shown in Fig. \ref{fig:distribution}. From the distortion MOS distributions, we can see that the TC distortion tends to have a less negative impact while the SBE distortions seem to cause more severe damage to the visual quality of DDHs. With closer inspections, we can observe that all types of motion exhibit similar MOS distributions, which indicates that the added distortions result in similar perceptual loss regardless of the motion types. Therefore the proposed DDH-QA database can provide useful guidelines for other types of DDH motion.

\section{Benchmark Experiment}
\subsection{Benchmark Competitors}
Since the DDH is usually perceived in the format of animated videos, several state-of-the-art video quality assessment (VQA) methods are employed for validation on the DDH-QA database. The FR methods include PSNR, and SSIM \cite{ssim}, which operate on the frames of DDH videos. The NR methods include BRISQUE \cite{brisque}, NIQE \cite{niqe}, VIIDEO \cite{mittal2015completely}, V-BLIINDS \cite{saad2014blind}, TLVQM \cite{korhonen2019two}, VIDEVAL \cite{tu2021ugc}, VSFA \cite{li2019quality}, RAPIQUE \cite{tu2021rapique}, SimpleVQA \cite{sun2022deep}, and FAST-VQA \cite{wu2022fast}. Additionally, BRISQUE, NIQE, VIIDEO, V-BLIINDS, TLVQM, and VIDEVAL are handcrafted-based methods while VSFA, RAPIQUE, SimpleVQA, and FAST-VQA are DNN-based methods. It's worth mentioning that we use the source codes provided by the authors and maintain the default setting parameters.

\subsection{Experimental Setup}
The 5-fold cross validation strategy is utilized to train and test the models. Specifically, we split the 10 groups of motion into 5 folds and each fold contains 2 groups of motion. 4 folds are used as the training sets while the left 1 fold is used as the testing set. Such procedure is repeated 5 times so that every fold has been employed as the testing set. The average performance is recorded as the final experimental results. Additionally, for methods that require no training, we simply operate them on the same testing sets and report the average performance.

Four mainstream consistency evaluation criteria are utilized to compare the correlation between the predicted scores and MOSs, which include Spearman Rank Correlation Coefficient (SRCC), Kendall’s Rank Correlation Coefficient (KRCC), Pearson Linear Correlation Coefficient (PLCC), and Root Mean Squared Error (RMSE).
An excellent model should obtain values of SRCC, KRCC, and PLCC close to 1, and the value of RMSE near 0.

\begin{table}[t]\small
\renewcommand\arraystretch{1.4}
\renewcommand\tabcolsep{4.2pt}
  \caption{Benchmark Performance on the DDH-QA database. Best in bold. }
  \begin{center}
  \begin{tabular}{c|c|cccc}
    \toprule
    Ref. &  Model & SRCC &  PLCC & KRCC & RMSE \\ \hline
    \multirow{2}{*}{FR} 
    &PSNR  &0.4308 &0.5458 &0.3114 &0.9013\\
    &SSIM  &0.5408 &0.6057 &0.3920 &0.8559\\ \hdashline
    \multirow{10}{*}{NR} 
    &BRISQUE &0.3664 &0.4011 &0.2568 &1.0067\\
    &NIQE &0.0923 &0.2489 &0.0748 &1.0418\\

    &VIIDEO &0.1219  &0.1829 &0.0732 &1.0740\\
    &V-BLIINDS &0.4807  &0.4936 &0.3424 &0.9564\\
    &TLVQM &0.2515  &0.2824 &0.1729 &1.0480\\
    &VIDEVAL &0.2218  &0.3470 &0.1622 &1.0246\\
    &VSFA &0.5406  &0.5708 &0.3858 &0.9657\\
    &RAPIQUE &0.1815  &0.2368 &0.1246 &1.0614\\
    &SimpleVQA &\textbf{0.7444}  & \textbf{0.7498} & \textbf{0.5452} & \textbf{0.7228}\\
    &FAST-VQA &0.5262  &0.5382 &0.3657 &1.0499\\

    \bottomrule
  \end{tabular}
  \end{center}
  \label{tab:performance}

\end{table}

\subsection{Performance Discussion}
The experimental results are shown in Table \ref{tab:performance}, from which we can make several useful conclusions. (a) PSNR and SSIM are the most widely used FR quality assessment metrics in compression and transmission systems. Although they achieve relatively better performance than the NR handcrafted-based methods, they are not effective to deal with practical DDH-QA tasks, which calls for better FR DDH-QA solutions. (b) For the NR-VQA methods, the DNN-based methods except RAPIQUE tend to yield better performance than the handcrafted-based methods. This is because the handcrafted-based methods are based on the natural scene statistics (NSS) prior knowledge, which is learned from natural videos. However, the rendered DDH videos are quite different from the natural videos in both content and distortions, which leads to the ineffectiveness of the handcrafted-based methods. (c) SimpleVQA achieves the highest performance among all the benchmark competitors and is significantly superior to the second-ranking method. We try to give some reasons. Besides using 2D-CNN for spatial feature extraction, SimpleVQA also utilizes 3D-CNN for motion feature extraction, which might be more capable of describing the quality representation of DDHs. To sum up, the existing quality assessment methods still have a long way to go before accurately predicting the visual quality levels of DDHs.
 
\section{Conclusion}
In this paper, we propose a large-scale dynamic digital human quality assessment database. One male and one female digital human models are selected as the reference. Then we degrade the reference models with both model-based and motion-based distortions. A total of 800 DDHs are generated and we render the DDHs into 2D animation videos for evaluation. Afterward, we carry out a subjective study to collect the subjective quality judgment for the distorted DDHs. Several state-of-the-art VQA methods are chosen for validation on the proposed DDH-QA database. A comprehensive performance discussion is made as well. We hope our work will draw more attention to the quality assessment of DDHs and inspire future research.   




\bibliographystyle{IEEEbib}
\bibliography{icme2022template}

\end{document}